\documentclass[10pt,twocolumn,letterpaper]{article}
\usepackage{iccv}
\usepackage{times}
\usepackage{epsfig}
\usepackage{graphicx}
\usepackage{amsmath}
\usepackage{amssymb}

\usepackage{booktabs}
\usepackage{url}
\usepackage{xcolor}
\usepackage{multirow}
\usepackage{bbding}
\usepackage{makecell}
\usepackage{colortbl} 
\usepackage{float}
\usepackage{blindtext}
\usepackage[accsupp]{axessibility}

\usepackage[noend]{algpseudocode}
\usepackage{algorithmicx,algorithm}

\usepackage[colorlinks=true,bookmarks=false]{hyperref}

\usepackage[capitalize]{cleveref}
\crefname{section}{Sec.}{Secs.}
\Crefname{section}{Section}{Sections}
\Crefname{table}{Table}{Tables}
\crefname{table}{Tab.}{Tabs.}

\iccvfinalcopy 

\def\iccvPaperID{6514} 

\ificcvfinal\fi

\begin{document}

\title{SLCA: Slow Learner with Classifier Alignment for Continual Learning \\ on a Pre-trained Model}


\author{
  Gengwei Zhang$^{1}$\thanks{Equal contribution.},\quad Liyuan Wang$^{2}$\footnotemark[1], \quad Guoliang Kang$^{3,4}$,\quad Ling Chen$^{1}$,\quad Yunchao Wei$^{5,6}$  \\
  $^{1}$ AAII, University of Technology Sydney \,\,
  $^{2}$ Tsinghua University \\  
  $^{3}$ Beihang University \,\,\,
  $^{4}$ Zhongguancun Laboratory\\
  $^{5}$ Institute of Information Science, Beijing Jiaotong University \\
  $^{6}$ Beijing Key Laboratory of Advanced Information Science and Network \\
  \small{\{zgwdavid, kgl.prml, wychao1987\}@gmail.com; wly19@mail.tsinghua.org.cn;
  ling.chen@uts.edu.au}
}

\maketitle
\ificcvfinal\thispagestyle{empty}\fi

\begin{abstract}
The goal of continual learning is to improve the performance of recognition models in learning sequentially arrived data. Although most existing works are established on the premise of learning from scratch, growing efforts have been devoted to incorporating the benefits of pre-training. However, how to adaptively exploit the pre-trained knowledge for each incremental task while maintaining its generalizability remains an open question. In this work, we present an extensive analysis for continual learning on a pre-trained model (CLPM), and attribute the key challenge to a progressive overfitting problem. Observing that selectively reducing the learning rate can almost resolve this issue in the representation layer, we propose a simple but extremely effective approach named Slow Learner with Classifier Alignment (SLCA), which further improves the classification layer by modeling the class-wise distributions and aligning the classification layers in a post-hoc fashion. Across a variety of scenarios, our proposal provides substantial improvements for CLPM (\textit{e.g.}, up to 49.76\%, 50.05\%, 44.69\% and 40.16\% on Split CIFAR-100, Split ImageNet-R, Split CUB-200 and Split Cars-196, respectively), and thus outperforms state-of-the-art approaches by a large margin.
Based on such a strong baseline, critical factors and promising directions are analyzed in-depth to facilitate subsequent research. Code has been made available at: \url{https://github.com/GengDavid/SLCA}.
\end{abstract}

\section{Introduction}
\label{sec:intro}
Continual learning aims to learn effectively from sequentially arrived data, behaving as if they were observed simultaneously. 
Current efforts are mainly based on the premise of learning from scratch, attempting to mitigate catastrophic forgetting \cite{mcclelland1995there} of previously-learned knowledge when adapting to each incremental task. 
However, the success of large-scale pre-training has revolutionized the training paradigm of deep neural networks. The pre-training stage brings both strong knowledge transfer and robustness to catastrophic forgetting for downstream continual learning \cite{wang2023comprehensive}, which tends to be more significant as the scale of pre-training increases \cite{ramasesh2021effect,mehta2021empirical}.
Therefore, continual learning on a pre-trained model (CLPM) turns out to be an emerging direction and receives growing attention.

\begin{figure}[t]
    \centering
    \includegraphics[width=0.95\linewidth]{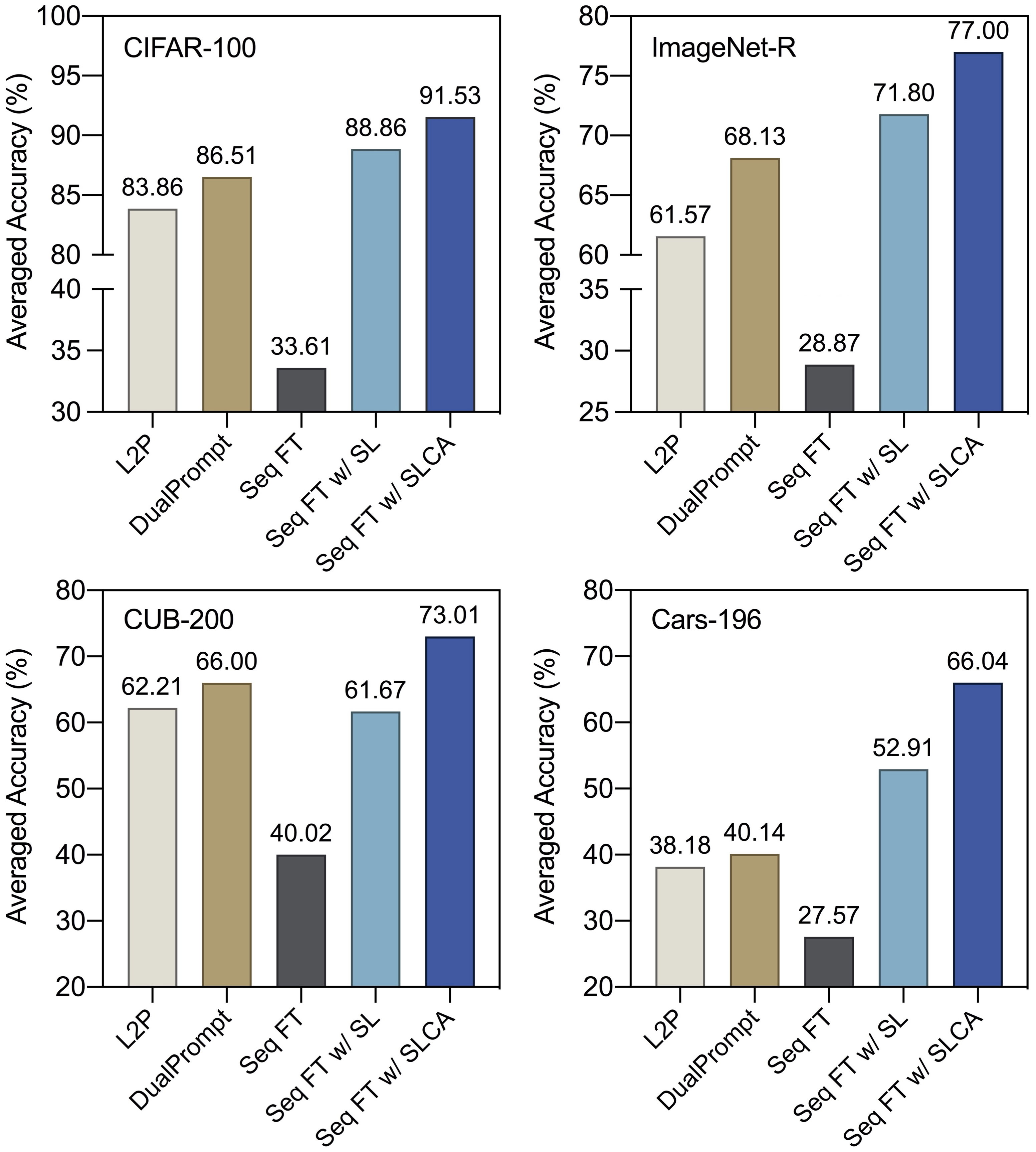}
    \caption{Performance of continual learning with supervised pre-training of ImageNet-21K. The proposed Slow Learner (SL) with Classifier Alignment (SLCA) enables sequential fine-tuning (Seq FT) to outperform prompt-based approaches such as L2P \cite{wang2022l2p} and DualPrompt \cite{wang2022dualprompt} by a large margin.}
    \label{img21k_sup}
\end{figure}


For CLPM, the pre-trained knowledge is usually expressed by the representation layer, adapted to a sequence of incremental tasks.
There are two major strategies to leverage the pre-trained knowledge \cite{wang2023comprehensive}:
(1) fine-tune the representations, or (2) keep the representations fixed while learning a few additional parameters (\eg, adaptor \cite{houlsby2019parameter}, prompt \cite{lester2021power}, instruction \cite{efrat2020turking}, etc.). 
Although (2) is becoming dominant in natural language processing (NLP) \cite{ke2022continual_review}, \emph{the choice of (1) and (2) remains an open question for continual learning in computer vision (CV)}.
The recently proposed prompt-based approaches, such as L2P \cite{wang2022l2p} and DualPrompt \cite{wang2022dualprompt}, followed the second strategy and reported to be \emph{far superior} to the traditional continual learning baselines of fine-tuning the representation layer.
On the other hand, since the large amount of pre-training data are typically unlabeled and may also arrive incrementally, it seems more reasonable to use self-supervised pre-training than supervised pre-training \cite{cossu2022continual,wang2023comprehensive}, also regarding that (upstream) continual learning in a self-supervised manner is generally more robust to catastrophic forgetting \cite{hu2021well,madaan2021rethinking,fini2022self}.

In this work, however, we present an extensive analysis to \emph{reconsider the current progress and technical route} for CLPM in CV.
Specifically, CLPM poses a critical challenge for continual learning that the pre-trained knowledge should be adaptively exploited for the current task while maintaining generalizability for future tasks. 
For traditional continual learning baselines that sequentially fine-tune (Seq FT) the entire model, a uniform learning rate is typically too large for the representation layer, leading to a \emph{progressive overfitting} problem where the pre-trained representations overfit to each incremental task and gradually lose generalizability. 
This is the major cause of their inferiority to the recent prompt-based approaches.
Such a phenomenon is further exacerbated by using the more realistic self-supervised pre-training, which generally requires larger updates of the pre-trained representations to adequately accommodate each incremental task.


To address the above challenges, we propose a simple but extremely effective approach named Slow Learner with Classifier Alignment (SLCA).
First, we observe that the progressive overfitting problem of the representation layer can be almost avoided by selectively reducing its learning rate (\textit{i.e.}, using a slow learner), which is sufficient to achieve a good trade-off between task specificity and generalizability in continually updating the pre-trained representations. 
Based on the proposed slow learner, we further improve the classification layer by modeling the class-wise distributions and aligning the classification predictions, so as to properly balance the relationships between different tasks.

Across a variety of continual learning benchmarks under supervised or self-supervised pre-training, our proposal provides substantial improvements for CLPM in CV and dramatically fills the gap of current progress from the upper bound performance. For example, the naive Seq FT is improved by \textbf{49.76\%}, \textbf{50.05\%}, \textbf{44.69\%} and \textbf{40.16\%} on Split CIFAR-100, Split ImageNet-R, Split CUB-200 and Split Cars-196, respectively, thus outperforming the SOTA approaches by a large margin (summarized in Fig.~\ref{img21k_sup}). On Split CIFAR-100 and Split ImageNet-R, the performance gap is only less than \textbf{2\%} for supervised pre-training and less than \textbf{4\%} for self-supervised pre-training.

Our contributions include three aspects: (1) We present an extensive analysis of continual learning on a pre-trained model (CLPM), and demonstrate that the progressive overfitting problem is the key challenge for traditional continual learning baselines.
(2) We propose a simple but effective approach to address this problem, which provides substantial improvements for CLPM and clearly outperforms the state-of-the-art approaches, serving as a strong baseline to re-evaluate the current progress and technical route.
(3) Our results further identify critical factors and promising directions for CLPM, such as pre-training paradigm and downstream granularity, so as to facilitate subsequent research.


\section{Related Work}
\label{sec:related}
Existing work on continual learning mainly focuses on sequential training of deep neural network(s) from scratch, seeking to effectively learn each new task without severely forgetting the old tasks. 
Representative strategies include regularization-based approaches \cite{kirkpatrick2017overcoming,aljundi2018memory,zenke2017continual,li2017learning,wang2021afec,dhar2019learning}, which preserve the old model and selectively stabilize changes of parameters or predictions; replay-based approaches \cite{wu2019large,prabhu2020gdumb,buzzega2020dark,wang2021memory,wang2021ordisco,wang2021triple}, which approximate and recover the previously-learned data distributions; architecture-based approaches \cite{serra2018overcoming,rusu2016progressive,wang2022coscl,yang2022continual}, which allocate dedicated parameter sub-spaces for each incremental task; etc. 

Recent works have increasingly explored the benefits of pre-training for continual learning.
For example, the representations obtained from supervised pre-training have been shown to facilitate not only knowledge transfer but also robustness to catastrophic forgetting for downstream continual learning \cite{ramasesh2021effect,mehta2021empirical,zhu2023ctp}.
Also, learning a large number of base classes in the initial training phase allows class-incremental learning with small adaptations \cite{wu2022class}.
Inspired by the techniques of using pre-trained knowledge in NLP, L2P \cite{wang2022l2p} employed an additional set of learnable parameters called ``prompts'' that dynamically instruct a pre-trained representation layer to learn incremental tasks. DualPrompt \cite{wang2022dualprompt} extended this idea by attaching complementary prompts to the pre-trained representation layer for learning task-invariant and task-specific instructions.
Such prompt-based approaches were reported to be far superior to traditional continual learning baselines, which potentially challenges the current paradigm of using pre-trained knowledge in CV.

Since the large amount of training samples required to construct a pre-trained model are typically unlabeled and may also arrive incrementally, self-supervised pre-training emerges as a more preferable choice than supervised pre-training. Several recent works discovered that continual learning in a self-supervised manner suffers from less catastrophic forgetting \cite{hu2021well,madaan2021rethinking,fini2022self}. Indeed, self-supervised paradigms have been shown to be better adapted to upstream continual learning \cite{cossu2022continual}.
However, the effectiveness of self-supervised pre-training for downstream continual learning remains to be investigated.

\section{Continual Learning on a Pre-trained Model}
\label{sec:method}
In this section, we introduce the problem formulation of continual learning on a pre-trained model (CLPM), perform an extensive analysis, and then present our approach.

\subsection{Problem Formulation}

Let's consider a neural network $M_{\theta}(\cdot) = h_{\theta_{cls}}(f_{\theta_{rps}}(\cdot))$ with parameters $\theta=\{\theta_{rps}, \theta_{cls}\}$ for classification tasks, consisting of a representation layer $f_{\theta_{rps}}(\cdot)$ that projects input images to feature representations, and a classification layer $h_{\theta_{cls}}(\cdot)$ that projects feature representations to output predictions. $\theta_{rps}$ is initialized on a pre-training dataset $D_{pt}$ in a supervised or self-supervised manner (labels are not necessary for the latter). 
Then, $M_{\theta}$ needs to learn a sequence of incremental tasks from their training sets $D_t, t=1,...,T$ and tries to perform well on their test sets.
Following previous works of CLPM in CV~\cite{wang2022l2p,wang2022dualprompt}, we mainly focus on the class-incremental scenario of continual learning \cite{vandeven2019three}. In details, $D_t = {\bigcup}_{c \in C_t} \{(x_{c,n}, y_{c,n})\}_{n=1}^{N_c} $ introduces a set of new classes $C_t$, where $N_c$ denotes the number of training samples $(x_{c,n}, y_{c,n})$ for class $c$, and all the classes ever seen are evaluated without task labels.

To achieve this aim, the network needs to (1) effectively transfer the pre-trained knowledge to each incremental task while maintaining its generalizability for future tasks, and (2) properly balance learning plasticity of new tasks with memory stability of old tasks. A naive baseline is to sequentially fine-tune the entire model $M_{\theta}$ on each $D_t$, where $f_{\theta_{rps}}$ and $h_{\theta_{cls}}$ are updated in a similar speed (\textit{i.e.}, using the same learning rate). However, due to the lack of $D_{pt}$ and $D_{1:t-1} = {\bigcup}_{i=1}^{t-1}D_i$, the performance of $M_{\theta}$ is severely limited by a \emph{progressive overfitting} problem in both aspects. Specifically, (1) the knowledge of $D_{pt}$ is interfered by $D_t$, as $\theta_{rps}$ is continually updated to accommodate incremental tasks while the generalizability obtained from the pre-training stage is progressively lost; and (2) the knowledge of $D_{1:t-1}$ is interfered by $D_t$, as $\theta_{cls}$ (and $\theta_{rps}$) catastrophically forgets the old tasks when learning each new task.

Most continual learning approaches \cite{kirkpatrick2017overcoming,li2017learning,prabhu2020gdumb,wu2019large,buzzega2020dark} are established on the premise of learning from scratch and focus on improving the second aspect. We call them ``traditional baselines''.
To address the first aspect, the newly proposed prompt-based approaches \cite{wang2022l2p,wang2022dualprompt} fixed the representation layer and employed an additional set of learnable parameters to instruct the pre-trained model, which reported significantly better performance for CLPM than other traditional baselines that update the representation layer.

\begin{figure}[t]
    \centering
    \includegraphics[width=0.80\linewidth]{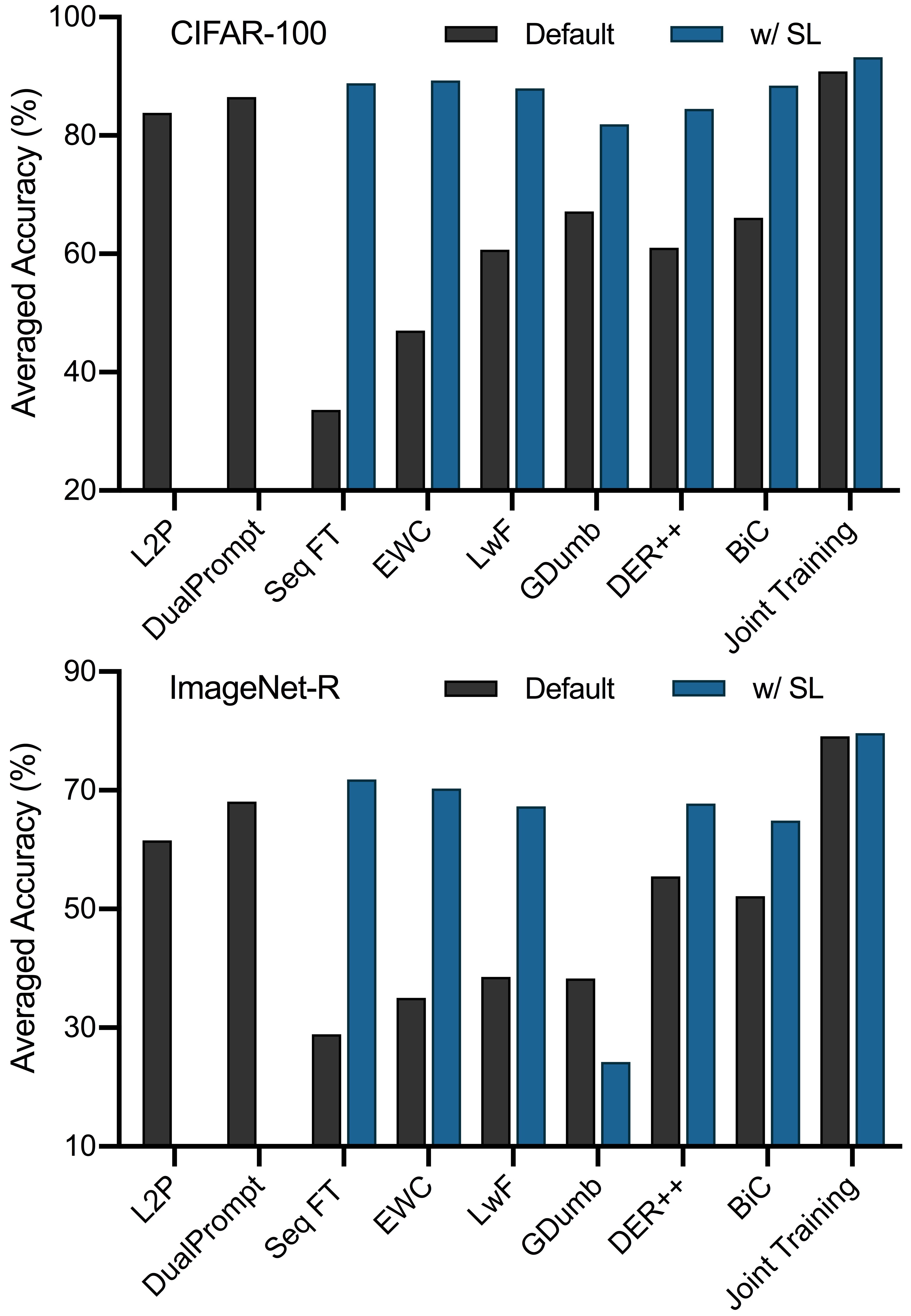}
    \caption{Slow Learner (SL) can greatly enhance continual learning performance on a pre-trained model. Here we adopt ImageNet-21K supervised pre-training for all baselines with default performance referenced from \cite{wang2022l2p,wang2022dualprompt},
    including prompt-based approaches (L2P \cite{wang2022l2p} and DualPrompt \cite{wang2022dualprompt}), regularization-based approaches (EWC \cite{kirkpatrick2017overcoming} and LwF \cite{li2017learning}), and replay-based approaches (GDumb \cite{prabhu2020gdumb}, DER++ \cite{buzzega2020dark} and BiC \cite{wu2019large}). }
    \label{img21k_sup_all}
\end{figure}

\begin{figure}[t]
    \centering
    \includegraphics[width=0.82\linewidth]{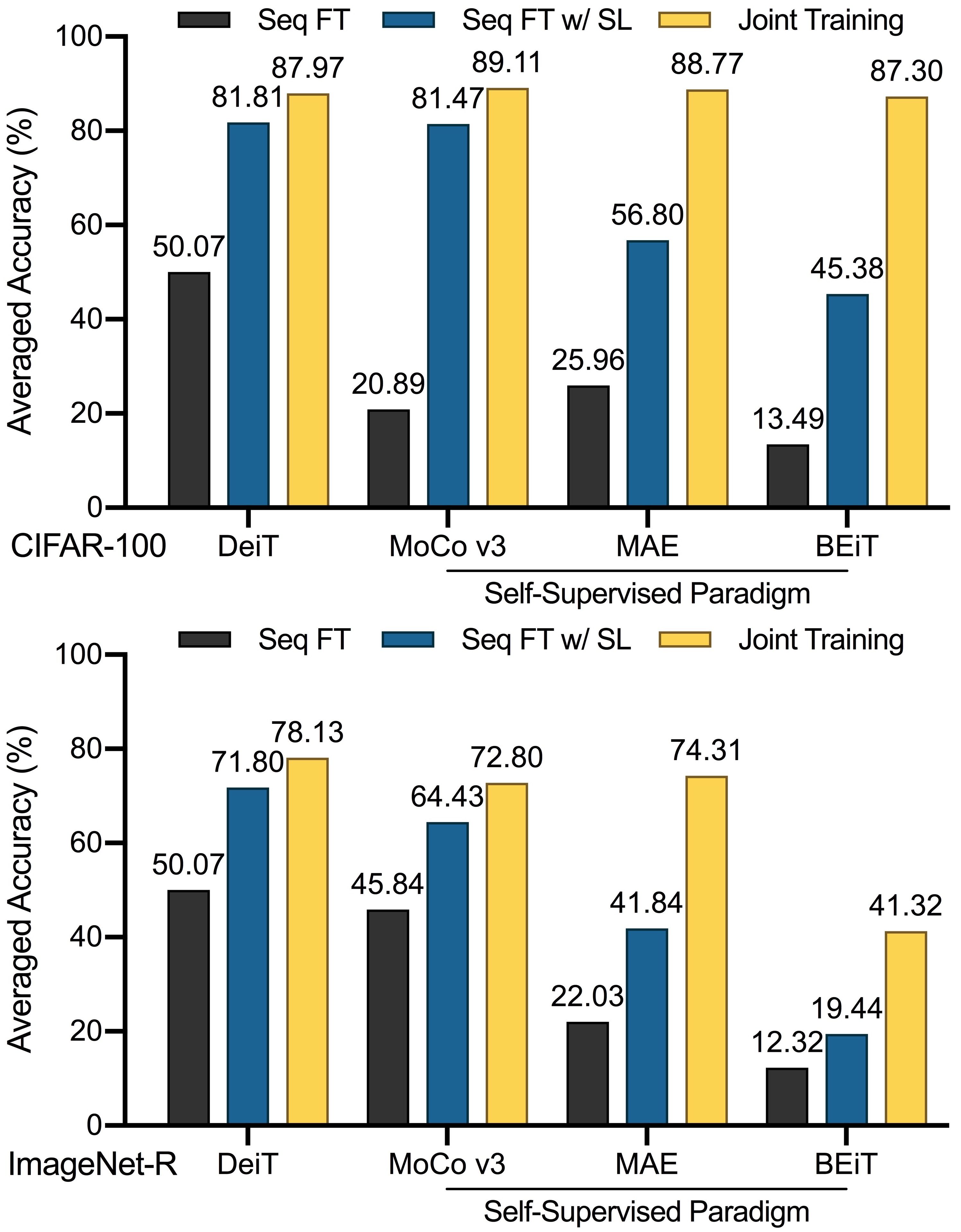}
    \caption{Comparison of pre-training paradigms on ImageNet-1K. DeiT \cite{touvron2021training} is a strong supervised method for (pre-)training vision transformer, while MoCo v3 \cite{chen2021empirical}, MAE \cite{he2022masked} and BEiT \cite{bao2021beit} are representative self-supervised methods. The pre-trained checkpoints are obtained from their official release. } 
    \label{img1k-cifar100-pt}
\end{figure}

\subsection{Slow Learner is (Almost) All You Need?}

\textbf{Implementation of Slow Learner:} However, we argue that the performance of traditional baselines reported in \cite{wang2022l2p,wang2022dualprompt} is severely limited by using a uniform learning rate (\textit{e.g.,} 0.005) for both $\theta_{rps}$ and $\theta_{cls}$. Through extensive experiments, 
we observe that using a much smaller learning rate (0.0001) for $\theta_{rps}$ and a slightly larger learning rate (0.01) for $\theta_{cls}$ can greatly enhance the performance of traditional baselines (Fig.~\ref{img21k_sup_all}).\footnote{A comprehensive analysis of the effect of learning rate can be found in the appendix.} The reported performance of sequential fine-tuning (Seq FT) is improved by more than 40\% for challenging continual learning benchmarks such as Split CIFAR-100 and Split ImageNet-R, respectively, thus clearly outperforming the prompt-based approaches such as L2P \cite{wang2022l2p} and DualPrompt \cite{wang2022dualprompt}.\footnote{Please note that the prompt-based approaches \cite{wang2022l2p,wang2022dualprompt} use a fixed representation layer, so our proposal is not applicable for them.} 
We call this simple but surprisingly effective strategy ``Slow Learner (SL)'', corresponding to slowing down the updating speed of the representation layer.
In contrast to previous works of using different learning rates for transfer learning \cite{guo2019spottune,zhang2020revisiting,he2019rethinking}, the benefit of our proposal is \emph{specific to continual learning}, as the upper bound performance (\textit{i.e.}, joint training) is only marginally improved by the SL while the performance gap between continual learning and joint training is greatly filled (\textit{e.g.}, only 4.36\% on Split CIFAR-100 and 7.80\% on Split ImageNet-R for Seq FT w/ SL).

\begin{figure}[t]
    \centering
    \includegraphics[width=0.90\linewidth]{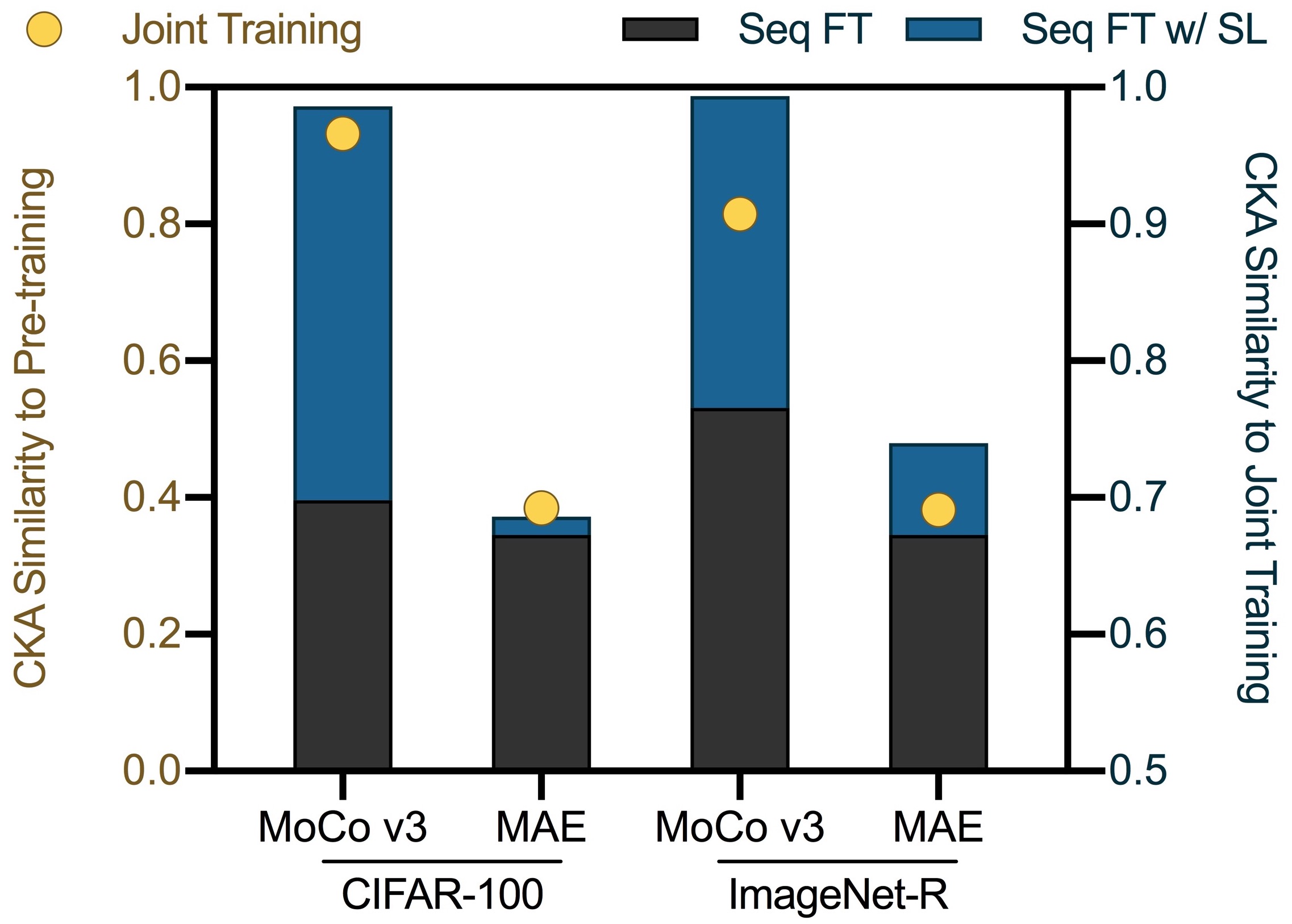}
    \caption{Similarity of the pre-trained representations (1) before and after joint training (left Y-axis, yellow dot), and (2) after joint training and after continual learning (right Y-axis, column).
    We adopt Centered Kernel Alignment (CKA) \cite{kornblith2019similarity} as the similarity metric. \emph{Best viewed in color.}
    }
    \label{ssl_cka_jt_seq}
\end{figure}

\textbf{Effect of Pre-training Paradigm:} We then evaluate the effect of pre-training paradigms (\textit{i.e.}, supervised or self-supervised) on downstream continual learning. 
Here we focus on ImageNet-1K as the pre-training dataset, since most self-supervised methods only release checkpoints on it.
Considering architectural consistency with previous works of CLPM \cite{wang2022l2p,wang2022dualprompt}, we select representative self-supervised methods (\textit{i.e.,} MoCo v3 \cite{chen2021empirical}, MAE \cite{he2022masked} and BEiT \cite{bao2021beit}) that release checkpoints on ViT-B/16 in our comparisons. We further compare DeiT \cite{touvron2021training}, a strong supervised method for (pre-)training vision transformer.
As shown in Fig.~\ref{img1k-cifar100-pt}, self-supervised pre-training, while more realistic regarding labeling requirements and upstream continual learning, typically results in a larger performance gap between Seq FT and joint training than supervised pre-training. 
Similarly, the use of SL can effectively reduce this gap. 
Interestingly, the performance of Seq FT w/ SL for MoCo v3 \cite{chen2021empirical} far exceeds that of the more recent MAE \cite{he2022masked}, although their joint training performance is comparable. This is possibly because the pre-trained representations of MoCo v3 \cite{chen2021empirical} require remarkably smaller updates to learn all tasks well (Fig.~\ref{ssl_cka_jt_seq}, left Y-axis), thus alleviating the progressive overfitting problem. Meanwhile, the use of SL allows MoCo v3 \cite{chen2021empirical} to learn representations much closer to that of the joint training (Fig.~\ref{ssl_cka_jt_seq}, right Y-axis).
The above results suggest a new direction for the design of self-supervised paradigms, \textit{i.e.}, how to effectively perform downstream continual learning and combine the advantages of SL.

\begin{figure*}[t]
    \centering
    \includegraphics[width=1\linewidth]{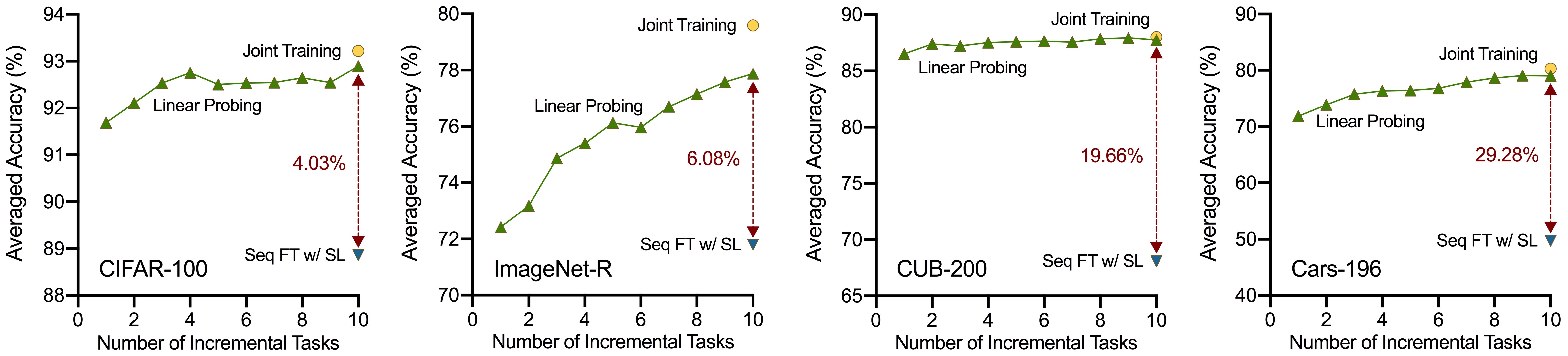}
    \caption{Linear probing results of Slow Learner. All experiments are based on ImageNet-21K supervised pre-training. We report the averaged accuracy of all classes in the corresponding benchmark dataset (\textit{e.g.}, a total of 100 classes in CIFAR-100 dataset). The dark red arrow represents the performance gap caused by a sub-optimal classification layer.
    }
    \label{img21k-cifar100-probe}
\end{figure*}

\textbf{Evaluation of Representation:} So, why is the Slow Learner such effective, and what accounts for the remaining performance gap? We perform a linear probing experiment \cite{he2022masked} to evaluate the performance of the representation layer. Specifically, after learning each incremental task (\textit{e.g.}, 10 classes per task for 10 tasks in Split CIFAR-100) via Seq FT w/ SL, we fix the representation layer and employ an extra classification layer, called a linear probe, to learn all classes of the corresponding benchmark dataset (\textit{e.g.}, a total of 100 classes in CIFAR-100 dataset). The performance of these linear probes is presented in Fig.~\ref{img21k-cifar100-probe}, which tends to grow with learning more tasks, indicating that the representation layer is accumulating knowledge for better adaptation. After learning all incremental tasks, it can be clearly seen that using the continually-learned representation layer to jointly train an extra classifier for all classes can almost reach the joint training performance of the entire model, and far outperform its counterpart with a continually-learned classifier (\textit{i.e.}, Seq FT w/ SL in Fig.~\ref{img21k-cifar100-probe}). Therefore, the proposed SL can almost address the problem of the representation layer, yet the classification layer remains sub-optimal. In particular, the problem of classification layer becomes more severe for fine-grained continual learning benchmarks such as Split CUB-200 and Split Cars-196.




\begin{figure}[t]
    \centering
    \includegraphics[width=1\linewidth]{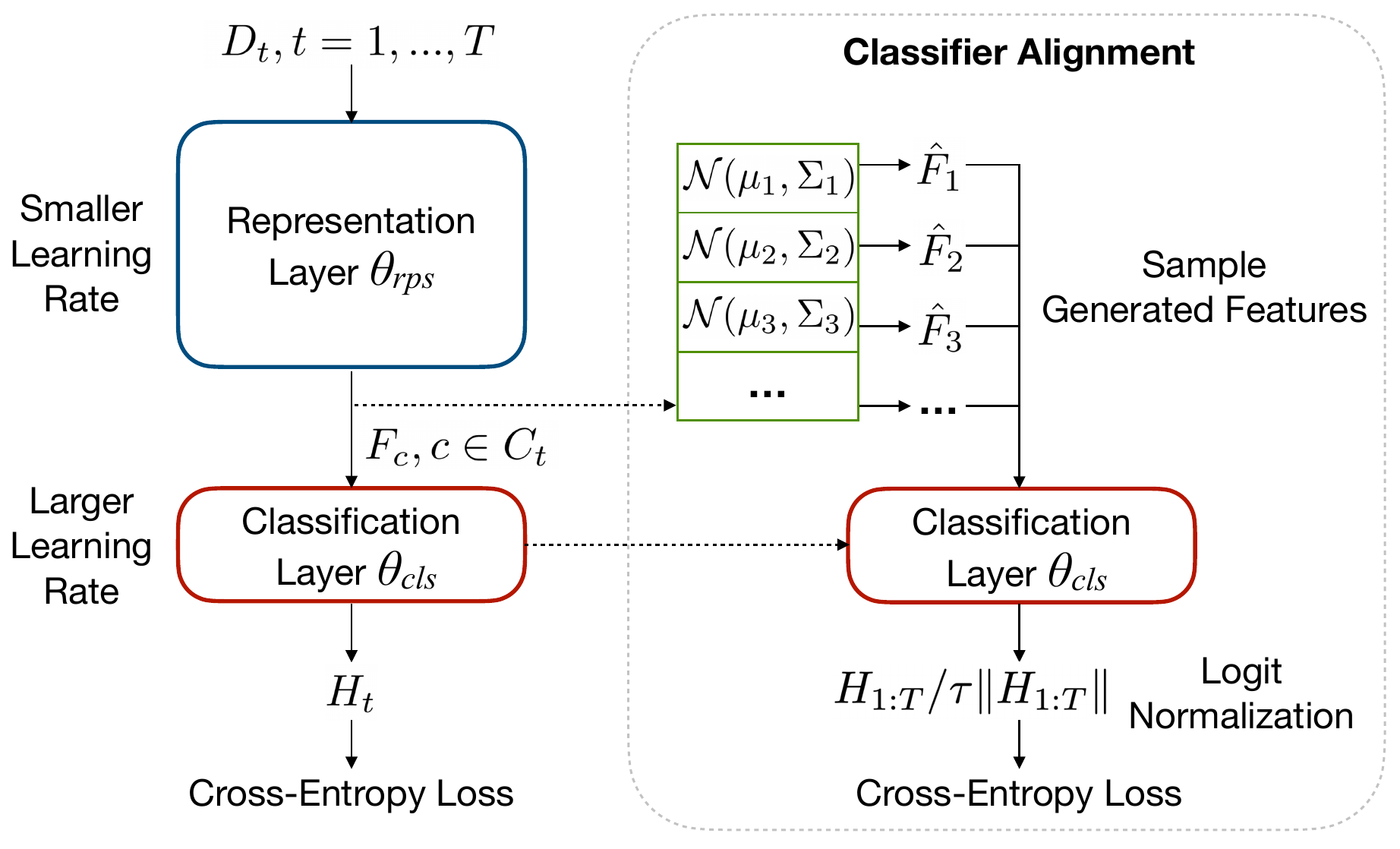}
    \caption{Slow Learner with Classifier Alignment (SLCA). $H_{t}$ is the logit of predicting the current training data in $D_t$.}
    \label{SLCA}
\end{figure}

\subsection{Slow Learner with Classifier Alignment}
To further improve the classification layer, we propose to align the statistics of previously-learned classes in a \textit{post-hoc} fashion (see Fig.~\ref{SLCA} and Algorithm~\ref{pesudocode}). Specifically, when learning each incremental task, only the parameters corresponding to the classes of the current task are trained together with the representation layer. After learning each task, we collect feature representations $F_c=[r_{c,1}, ... , r_{c,N_c}]$ for each class $c \in C_t$ of the current task, where $r_{c,n} = f_{\theta_{rps}}(x_{c,n})$ and $N_c$ denotes its amount. 
Instead of saving the extracted features $F_{c}$ of training samples, CA preserves their mean $\mu_c \in \mathbb{R}^{d}$ and covariance $\Sigma_c \in \mathbb{R}^{d \times d}$ for each class $c$ ($d$ denotes the feature dimension). Since the use of pre-training provides well-distributed representations, each class tends to be single-peaked and can be naturally modelled as a Gaussian $\mathcal{N}(\mu_c, \Sigma_c)$. 

Whenever the model needs to be evaluated, the classification layers are further aligned as follows.
We first sample generated features $\hat{F}_c=[\hat{r}_{c,1}, ... , \hat{r}_{c,S_c}]$ from the distribution $\mathcal{N}(\mu_c, \Sigma_c)$ of each class $c \in C_{1:T}$ ever seen in $C_{1:T} = \bigcup_{i=1}^{T} C_i$, where $S_c$ is the amount of generated features for each class ($S_{c}=256$ in our experiments), and the task amount $T$ can be any positive integer without being known in advance. 
Next, we apply a widely-used cross-entropy loss to adjust the classification layer $h_{\theta_{cls}}$ by feeding $\hat{F}_{1:T} = [\hat{F}_1,...,\hat{F}_{[C_{1:T}]}]$ as the input of the classification layer, where $[C_{1:T}]$ denotes the number of classes in $C_{1:T}$.

\begin{algorithm}[ht]
\caption{\small Slow Learner with Classifier Alignment (SLCA)}
\label{pesudocode}
{\bf Inputs:} Pre-training dataset $D_{pt}$; training dataset $D_t$ for task $t=1,...,T$; network $M_{\theta}(\cdot) = h_{\theta_{cls}}(f_{\theta_{rps}}(\cdot))$ with parameters $\theta=\{\theta_{rps}, \theta_{cls}\}$; learning rates $\alpha$ for $\theta_{rps}$ and $\beta$ for $\theta_{cls}$ ($\alpha < \beta$); temperature hyperparameter $\tau$.
\\
{\bf Initialization:} 
Initialize $\theta_{rps}$ by pre-training on $D_{pt}$; initialize $\theta_{cls}$ randomly.
\begin{algorithmic}[1]
\State \textit{\# sequential tasks.} 
\For{task $t=1,...,T$} 
 \State \textit{\# different learning rates for $\theta_{rps}$ and $\theta_{cls}$.}
 \While{not converged} 
 \State 
 Train $M_{\theta}$ with cross-entropy loss on $D_t$. 
 \State \textbf{end while}
 \EndWhile
  \State Collect $F_c=[r_{c,1}, ... , r_{c,N_c}]$ for $c \in C_t$.
  \State Save mean $\mu_c$ and covariance $\Sigma_c$ of $F_c$ for $c \in C_t$.
\State \textbf{end for}
\EndFor
  \State \textit{\# classifier alignment.}
  \State Sample $\hat{F}_c$ from $\mathcal{N}(\mu_c, \Sigma_c)$ for $c \in C_{1:T}$.
 \While{not converged} 
 \State Compute logit $H_{1:T}$ and its magnitude $\| H_{1:T} \|$. 
 \State {Train $h_{\theta_{cls}}$ with normalized logit in Eqn.~\ref{eqn_modifed_ce}.}
  \State \textbf{end while}
 \EndWhile
\end{algorithmic}
\end{algorithm}

\begin{table*}[ht]
    \centering
    \footnotesize{
    \begin{tabular}{l|c|c||c|c||c|c}
    \hline
        \multirow{2}{*}{Method} & \multirow{2}{*}{Memory-Free} & \multirow{2}{*}{Pre-trained} & \multicolumn{2}{c||}{Split CIFAR-100} & \multicolumn{2}{c}{Split ImageNet-R} \\ \cline{4-7}
        & & & Last-Acc (\%)& Inc-Acc (\%)& Last-Acc (\%)& Inc-Acc (\%)\\ \hline
        Joint-Training & - & IN21K-Sup & 93.22\tiny{$\pm 0.16$} & - & 79.60\tiny{$\pm 0.87$} & -  \\ \hline
        GDumb \cite{prabhu2020gdumb}&  & IN21K-Sup & 81.92\tiny{$\pm 0.15$} & 89.46\tiny{$\pm 0.94$} & 24.23\tiny{$\pm 0.35$} & 43.48\tiny{$\pm 0.49$}  \\ 
        DER++ \cite{buzzega2020dark}& & IN21K-Sup & 84.50\tiny{$\pm 1.67$} & 91.49\tiny{$\pm 0.61$} & 67.75\tiny{$\pm 0.93$} & 78.13\tiny{$\pm 1.14$} \\ 
        BiC \cite{wu2019large}& & IN21K-Sup & 88.45\tiny{$\pm 0.57$} & 93.37\tiny{$\pm 0.32$} & 64.89\tiny{$\pm 0.80$} & 73.66\tiny{$\pm 1.61$}  \\ 
        L2P \cite{wang2022l2p}& $\checkmark$ & IN21K-Sup & 82.76\tiny{$\pm 1.17$} & 88.48\tiny{$\pm 0.83$} & 66.49\tiny{$\pm 0.40$} & 72.83\tiny{$\pm 0.56$} \\ 
        DualPrompt \cite{wang2022dualprompt}& $\checkmark$ & IN21K-Sup & 85.56\tiny{$\pm 0.33$} & 90.33\tiny{$\pm 0.33$} & 68.50\tiny{$\pm 0.52$} & 72.59\tiny{$\pm 0.24$}  \\ 
        EWC \cite{kirkpatrick2017overcoming}& $\checkmark$ & IN21K-Sup & 89.30\tiny{$\pm 0.23$} & 92.31\tiny{$\pm 1.66$}  & 70.27\tiny{$\pm 1.99$} & 76.27\tiny{$\pm 2.13$}  \\ 
        LwF \cite{li2017learning}& $\checkmark$ & IN21K-Sup & 87.99\tiny{$\pm 0.05$} & 92.13\tiny{$\pm 1.16$}  & 67.29\tiny{$\pm 1.67$} & 74.47\tiny{$\pm 1.48$}  \\ 
        Seq FT & $\checkmark$ & IN21K-Sup & 88.86\tiny{$\pm 0.83$} & 92.01\tiny{$\pm 1.71$} & 71.80\tiny{$\pm 1.45$} & 76.84\tiny{$\pm 1.26$}  \\ 
         \rowcolor{black!15}
        SLCA (Ours) & $\checkmark$ & IN21K-Sup & \textbf{91.53}\tiny{$\pm 0.28$} & \textbf{94.09}\tiny{$\pm 0.87$} & \textbf{77.00}\tiny{$\pm 0.33$} & \textbf{81.17}\tiny{$\pm 0.64$} \\ 
        \hline \hline
        Joint-Training & - & IN1K-Self & 89.11\tiny{$\pm 0.06$} & - & 72.80\tiny{$\pm 0.23$} & -  \\ \hline 
        GDumb \cite{prabhu2020gdumb}&  & IN1K-Self & 69.72\tiny{$\pm 0.20$} & 80.95\tiny{$\pm 1.19$} & 28.24\tiny{$\pm 0.58$} & 43.64\tiny{$\pm 1.05$}  \\ 
        DER++ \cite{buzzega2020dark}&  & IN1K-Self & 63.64\tiny{$\pm 1.30$} & 79.55\tiny{$\pm 0.87$} & 53.11\tiny{$\pm 0.44$} & 65.10\tiny{$\pm 0.91$} \\ 
        BiC \cite{wu2019large}& & IN1K-Self & 80.57\tiny{$\pm 0.86$} & 89.39\tiny{$\pm 0.33$} & 57.36\tiny{$\pm 2.68$} & 68.07\tiny{$\pm 0.22$} \\ 
        EWC \cite{kirkpatrick2017overcoming}& $\checkmark$ & IN1K-Self & 81.62\tiny{$\pm 0.34$} & 87.56\tiny{$\pm 0.97$} & 64.50\tiny{$\pm 0.36$} & 70.37\tiny{$\pm 0.41$} \\ 
        LwF \cite{li2017learning}& $\checkmark$ & IN1K-Self & 77.94\tiny{$\pm 1.00$} & 86.90\tiny{$\pm 0.90$} & 60.74\tiny{$\pm 0.30$} & 68.55\tiny{$\pm 0.65$} \\ 
        Seq FT & $\checkmark$ & IN1K-Self & 81.47\tiny{$\pm 0.55$} & 87.55\tiny{$\pm 0.95$} & 64.43\tiny{$\pm 0.44$} & 70.48\tiny{$\pm 0.54$} \\
         \rowcolor{black!15}
        SLCA (Ours) & $\checkmark$ & IN1K-Self & \textbf{85.27}\tiny{$\pm 0.08$} & \textbf{89.51}\tiny{$\pm 1.04$} & \textbf{68.07}\tiny{$\pm 0.21$} & \textbf{73.04}\tiny{$\pm 0.56$} \\ \hline
    \end{tabular}
    }
    \vspace{+0.1cm}
    \caption{Experimental results for continual learning on Split CIFAR-100 and Split ImageNet-R. IN21K-Sup: supervised pre-training on ImageNet-21K. IN1K-Self: self-supervised pre-training on ImageNet-1K with MoCo v3 \cite{chen2021empirical}.
    All other fine-tuning based methods are reproduced according to their officially-released codes with the proposed \textbf{Slow Learner} implemented.}
\label{table:cifar+inr}
\end{table*}

However, a prolonged training of the classification layer can lead to an overconfidence issue, which potentially impairs generalizability to the test set(s).
To overcome this, we draw inspirations from out-of-distribution (OOD) detection \cite{wei2022mitigating} and normalize the magnitude of network outputs when computing the cross-entropy. Let $H_{1:T} = h_{\theta_{cls}}(\hat{F}_{1:T}) = [h_{\theta_{cls}}(\hat{F}_{1}),..., h_{\theta_{cls}}(\hat{F}_{[C_{1:T}]})] := [l_{1},..., l_{[C_{1:T}]}]$ denote the logit (\textit{i.e.}, pre-softmax output) of $\hat{F}_{1:T}$, which can be re-written as the product of two components:
$H_{1:T} = \| H_{1:T} \| \cdot \vec{H}_{1:T},$ 
where $\| \cdot \|$ denotes $L_2$-norm. $\| H_{1:T} \| = \sqrt{\sum_{c \in C_{1:T}} \| l_{c} \|^2} $ represents the magnitude of $H_{1:T}$, and $\vec{H}_{1:T}$ represents its direction. Then we adopt a modified cross-entropy loss with logit normalization to perform classifier alignment:
\begin{equation}
\mathcal{L}(\theta_{cls}; \hat{F}_{1:T} ) = - \log \frac{e^{{l_y} / (\tau \|H_{1:T}\|)}}{\sum_{c \in C_{1:T}} e^{{l_{c}} / (\tau \|H_{1:T}\|)}},
\label{eqn_modifed_ce}
\end{equation}
where $l_y$ denotes the $y$-th element of $H_{1:T}$ corresponding to the ground-truth label $y$. $\tau$ is a temperature hyperparameter. 
The intuition is that normalizing $H_{1:T}$ with an input-dependent constant $\tau \|H_{1:T}\|$ will not change the result of softmax prediction $\arg\max_{c \in C_{1:T}}(l_{c})$, while forcing the magnitude $\|H_{1:T}\|$ before softmax becomes $\frac{1}{\tau}$ can make the criterion only adjust the direction $\vec{H}_{1:T}$ \cite{wei2022mitigating}. Therefore, the normalization in Eqn.~\ref{eqn_modifed_ce} can alleviate the overconfidence issue in classifier alignment.
In practice, we observe that the temperature hyperparameter is not sensitive and empirically find $\tau = 0.1$ to be a reasonable choice.



\begin{table*}[ht]
    \centering
    \footnotesize{
    \begin{tabular}{l|c|c||c|c||c|c}
    \hline
        \multirow{2}{*}{Method} & \multirow{2}{*}{Memory-Free} & \multirow{2}{*}{Pre-trained} & \multicolumn{2}{c||}{Split CUB-200} & \multicolumn{2}{c}{Split Cars-196} \\ \cline{4-7}
        & & & Last-Acc (\%) & Inc-Acc (\%)& Last-Acc (\%)& Inc-Acc (\%)\\ \hline
        Joint-Training & - & IN21K-Sup & 88.00\tiny{$\pm 0.34$} & -  & 80.31\tiny{$\pm 0.13$} & -  \\ 
        \hline
        GDumb \cite{prabhu2020gdumb} &  & IN21K-Sup & 61.80\tiny{$\pm 0.77$} & 79.76\tiny{$\pm 0.18$} & 25.20\tiny{$\pm 0.84$} & 49.48\tiny{$\pm 0.74$} \\
        DER++ \cite{buzzega2020dark}&  & IN21K-Sup & 77.42\tiny{$\pm 0.71$} & 87.61\tiny{$\pm 0.09$} & 60.41\tiny{$\pm 1.76$} & 75.04\tiny{$\pm 0.57$} \\  
        BiC \cite{wu2019large}& & IN21K-Sup & 81.91\tiny{$\pm 2.59$} & 89.29\tiny{$\pm 1.57$} & 63.10\tiny{$\pm 5.71$} & 73.75\tiny{$\pm 2.37$} \\ 
        L2P \cite{wang2022l2p}& \checkmark & IN21K-Sup & 62.21\tiny{$\pm 1.92$} & 73.83\tiny{$\pm 1.67$} & 38.18\tiny{$\pm 2.33$} & 51.79\tiny{$\pm 4.19$} \\ 
        DualPrompt \cite{wang2022dualprompt}& \checkmark & IN21K-Sup & 66.00\tiny{$\pm 0.57$} & 77.92\tiny{$\pm 0.50$} & 40.14\tiny{$\pm 2.36$} & 56.74\tiny{$\pm 1.78$} \\ 
        EWC \cite{kirkpatrick2017overcoming}& \checkmark & IN21K-Sup & 68.32\tiny{$\pm 2.64$} & 79.95\tiny{$\pm 2.28$} & 52.50\tiny{$\pm 3.18$} & 64.01\tiny{$\pm 3.25$}  \\ 
        LwF \cite{li2017learning}& \checkmark & IN21K-Sup & 69.75\tiny{$\pm 1.37$} & 80.45\tiny{$\pm 2.08$} & 49.94\tiny{$\pm 3.24$} & 63.28\tiny{$\pm 1.11$}  \\
        Seq FT & \checkmark & IN21K-Sup & 68.07\tiny{$\pm 1.09$} & 79.04\tiny{$\pm 1.69$} & 49.74\tiny{$\pm 1.25$} & 62.83\tiny{$\pm 2.16$} \\ 
         \rowcolor{black!15}
        SLCA (Ours) & \checkmark & IN21K-Sup & \textbf{84.71}\tiny{$\pm 0.40$} & \textbf{90.94}\tiny{$\pm 0.68$} & \textbf{67.73}\tiny{$\pm 0.85$} & \textbf{76.93}\tiny{$\pm 1.21$} \\ 
        \hline \hline
        Joint-Training & - & IN1K-Self & 79.55\tiny{$\pm 0.04$} & - & 74.52\tiny{$\pm 0.09$} & -  \\ 
        \hline
        GDumb \cite{prabhu2020gdumb}&& IN1K-Self & 45.29\tiny{$\pm 0.97$} & 66.86\tiny{$\pm 0.63$} & 20.95\tiny{$\pm 0.42$} & 45.40\tiny{$\pm 0.66$} \\ 
        DER++ \cite{buzzega2020dark}& & IN1K-Self  & 61.47\tiny{$\pm 0.32$} & 77.15\tiny{$\pm 0.61$} & 50.64\tiny{$\pm 0.70$} & 67.64\tiny{$\pm 0.45$} \\ 
        BiC \cite{wu2019large}& & IN1K-Self & \textbf{74.39}\tiny{$\pm 1.12$} & \textbf{82.13}\tiny{$\pm 0.33$} & 65.57\tiny{$\pm 0.93$} & \textbf{73.95}\tiny{$\pm 0.29$} \\ 
        EWC \cite{kirkpatrick2017overcoming}& \checkmark & IN1K-Self & 61.36\tiny{$\pm 1.43$} & 72.84\tiny{$\pm 2.18$}  & 53.16\tiny{$\pm 1.45$} & 63.61\tiny{$\pm 1.06$} \\ 
        LwF \cite{li2017learning}& \checkmark & IN1K-Self & 61.66\tiny{$\pm 1.95$} & 73.90\tiny{$\pm 1.91$}  & 52.45\tiny{$\pm 0.48$} & 63.87\tiny{$\pm 0.31$}  \\ 
        Seq FT & \checkmark & IN1K-Self & 61.67\tiny{$\pm 1.37$} & 73.25\tiny{$\pm 1.83$} & 52.91\tiny{$\pm 1.61$} & 63.32\tiny{$\pm 1.31$} \\ 
         \rowcolor{black!15}
        SLCA (Ours) & \checkmark & IN1K-Self & 73.01\tiny{$\pm 0.16$} & \textbf{82.13}\tiny{$\pm 0.34$} &  \textbf{66.04}\tiny{$\pm 0.08$} & 72.59\tiny{$\pm 0.04$} \\ 
        \hline
    \end{tabular}
    }
    \vspace{+0.1cm}
    \caption{Experimental results for continual learning on Split CUB-200 and Split Cars-196. IN21K-Sup: supervised pre-training on ImageNet-21K. IN1K-Self: self-supervised pre-training on ImageNet-1K with MoCo v3 \cite{chen2021empirical}. The results of all fine-tuning based baselines are reproduced according to their officially-released codes with the proposed \textbf{Slow Learner} implemented.}
\label{table:cub+cars}
\end{table*}

\section{Experiments}
\label{sec:exp}
In this section, we first briefly describe the experimental setups, and then present the experimental results. 

\subsection{Experimental Setups}

\textbf{Benchmark:} 
Following L2P~\cite{wang2022l2p} and DualPrompt~\cite{wang2022dualprompt}, we adopt pre-training from ImageNet-21K dataset \cite{ridnik2021imagenet21k}, also known as the full ImageNet \cite{deng2009imagenet} consisting of 14,197,122 images with 21,841 classes. We also consider pre-training from ImageNet-1K dataset \cite{krizhevsky2012imagenet}, a subset of ImageNet-21K introduced for the ILSVRC2012 visual recognition challenge, consisting of 1000-class images.

To evaluate the performance of downstream continual learning, we consider four representative benchmark datasets and randomly split each of them into 10 disjoint tasks:
The first two follow previous works \cite{wang2022l2p,wang2022dualprompt} and are relatively coarse-grained in terms of classification, while the last two are relatively fine-grained.
Specifically, CIFAR-100 dataset \cite{krizhevsky2009learning} consists of 100-class natural images with 500 training samples per class. ImageNet-R dataset \cite{hendrycks2021many} contains 200-class images, spliting into 24,000 and 6,000 images for training and testing (similar ratio for each class), respectively. Note that although the image categories of ImageNet-R are overlapped with ImageNet-21K, all images are out-of-distribution samples for the pre-train dataset, \textit{i.e.}, hard examples from ImageNet or newly collected data of different styles. It requires considerable adaptations of the pre-trained model, therefore serving as a challenging benchmark for continual learning. CUB-200 dataset \cite{wah2011caltech} includes 200-class bird images with around 60 images per class, 30 of which are used for training and the rest for testing. Cars-196 dataset \cite{krause20133d} includes 196 types of car images, split into 8,144 and 8,040 images for training and testing (similar ratio for each class), respectively.
We present the average accuracy of all classes after learning the last task, denoted as Last-Acc (equivalent to ``Avg. Acc'' in \cite{wang2022l2p,wang2022dualprompt}). We also compute the average accuracy of the classes ever seen after learning each incremental task and then present their average, denoted as Inc-Acc.


\textbf{Implementation:}
Following previous works \cite{wang2022l2p,wang2022dualprompt}, we adopt a pre-trained ViT-B/16 backbone for all baselines. In addition to supervised pre-training, we consider representative self-supervised paradigms that provide pre-trained checkpoints on ViT-B/16, \textit{i.e.,} MoCo v3 \cite{chen2021empirical}, BEiT \cite{bao2021beit} and MAE \cite{he2022masked}. For continual learning of downstream tasks, we follow the previous implementation that employs an Adam optimizer for L2P \cite{wang2022l2p} and DualPrompt \cite{wang2022dualprompt} while a SGD optimizer for other baselines, with the same batch size of 128. Our Slow Learner adopts a learning rate of 0.0001 for the representation layer and 0.01 for the classification layer, different from \cite{wang2022l2p,wang2022dualprompt} using 0.005 for the entire model. 

\textbf{Baseline:}
We adopt joint training 
as the upper bound performance and consider continual learning baselines with or without replaying old training samples. 
As for the former, a memory buffer of 1000 images is maintained, and we evaluate three representative replay-based approaches such as BiC \cite{wu2019large}, GDumb \cite{prabhu2020gdumb} and DER++ \cite{buzzega2020dark}. As for the latter, we evaluate representative regularization-based approaches such as EWC \cite{kirkpatrick2017overcoming} and LwF \cite{li2017learning}, and prompt-based approaches such as L2P \cite{wang2022l2p} and DualPrompt \cite{wang2022dualprompt}. 
Note that sequential fine-tuning usually serves as the lower bound performance of continual learning, but we observe that simply adjusting the learning rate (\textit{i.e.}, using the Slow Learner) makes it a surprisingly strong baseline.

\subsection{Experimental Results}

\textbf{Overall Performance:} All baseline approaches in Table~\ref{table:cifar+inr}, \ref{table:cub+cars} are trained with our \textbf{Slow Learner} for a fair comparison, except the prompt-based approaches that keep the representation layer fixed.
For continual learning of relatively coarse-grained classification tasks, such as Split CIFAR-100 and Split ImageNet-R in Table~\ref{table:cifar+inr} (also shown in Fig.~\ref{img21k_sup_all}), the SL can substantially enhance the performance of continual learning. With the help of Classifier Alignment (CA) and its Logit Normalization (LN), our approach clearly outperforms L2P \cite{wang2022l2p} and DualPrompt \cite{wang2022dualprompt}, and almost reach the joint training upper bound (the performance gap is less than \textbf{2\%} for supervised pre-training and \textbf{4\%} for self-supervised pre-training). 
 As for fine-grained continual learning benchmarks in Table~\ref{table:cub+cars}, SLCA achieves the strongest performance for supervised pre-training, while for self-supervised pre-training the performance is comparable or slightly inferior to the SL version of BiC \cite{wu2019large}, which additionally employs some old training samples. Please note that the presented implementation of SLCA can be seen as adding CA+LN together with SL to the simplest sequential fine-tuning (Seq FT) baseline. In fact, the proposed SLCA can be naturally plug-and-play with other continual learning approaches.
We leave it as a further work. 

It is worth noting that different replay-based approaches (w/ SL) behave differently in CLPM. 
In general, BiC \cite{wu2019large}, which updated the entire model with the old training samples in a similar way to learning the new ones and added a bias correction layer to mitigate imbalance between old and new classes, receives the most substantial improvements.
While GDumb \cite{prabhu2020gdumb}, which simply used the old training samples to train a new model from scratch at test time, has difficulty to adapt the pre-trained model with limited training samples and thus performs the worst. 

\newcommand{\tabincell}[2]{\begin{tabular}{@{}#1@{}}#2\end{tabular}}
\begin{table*}[ht]
	\centering
    \footnotesize{
 { 
	\begin{tabular}{c|l|c|c|c|c|c}
	 \hline
       & Method & Pre-trained & Split CIFAR-100 & Split ImageNet-R & Split CUB-200 & Split Cars-196 \\
        \hline
       \multirow{2}*{\tabincell{c}{Baseline}}
       &Seq FT $^{\dag}$&IN21K-Sup & 41.77\tiny{$\pm 13.8$} & 26.95\tiny{$\pm 11.8$} & 40.02\tiny{$\pm 1.08$} & 27.57\tiny{$\pm 1.79$} \\ 
       &w/ Fixed $\theta_{rps}$ &IN21K-Sup & 63.75\tiny{$\pm 0.67$}  & 34.64\tiny{$\pm 14.3$} & 60.44\tiny{$\pm 1.80$} & 24.51\tiny{$\pm 6.90$} \\
       \hline
      \multirow{5}*{\tabincell{c}{Ours}}
       &w/ SL &IN21K-Sup &88.86\tiny{$\pm 0.83$} &71.80\tiny{$\pm 1.45$} &68.07\tiny{$\pm 1.09$} &49.74\tiny{$\pm 1.25$} \\        
       &w/ Fixed $\theta_{rps}$+CA & IN21K-Sup & 75.64\tiny{$\pm 0.26$} & 50.73\tiny{$\pm 0.21$} & 82.71\tiny{$\pm 0.14$} & 54.45\tiny{$\pm 0.16$}   \\ 
       &w/ Fixed $\theta_{rps}$+CA+LN & IN21K-Sup & 75.62\tiny{$\pm 0.21$} & 51.83\tiny{$\pm 0.34$} & 83.65\tiny{$\pm 0.18$} & 53.43\tiny{$\pm 0.09$} \\   
       &w/ SL+CA &IN21K-Sup & 90.70\tiny{$\pm 0.52$} & 74.41\tiny{$\pm 0.51$} & 83.20\tiny{$\pm 0.19$} & \textbf{67.90}\tiny{$\pm 0.53$}  \\
       &w/ SL+CA+LN &IN21K-Sup &\textbf{91.53}\tiny{$\pm 0.28$}  &\textbf{77.00}\tiny{$\pm 0.33$}  & \textbf{84.71}\tiny{$\pm 0.40$}  & 67.73\tiny{$\pm 0.85$} \\ 
       \hline
       \hline
        \multirow{2}*{\tabincell{c}{Baseline}}  
       &Seq FT &IN1K-Self & 27.99\tiny{$\pm 5.16$} & 45.84\tiny{$\pm 4.19$} & 45.35\tiny{$\pm 1.38$} &  35.96\tiny{$\pm 2.04$} \\ 
       &w/ Fixed $\theta_{rps}$ &IN1K-Self & 77.30\tiny{$\pm 0.56$} & 51.97\tiny{$\pm 0.17$} & 55.54\tiny{$\pm 1.55$} & 43.16\tiny{$\pm 0.12$} \\ 
       \hline
      \multirow{5}*{\tabincell{c}{Ours}}
       &w/ SL &IN1K-Self &81.47\tiny{$\pm 0.55$} &64.43\tiny{$\pm 0.44$} &61.67\tiny{$\pm 1.37$} &52.91\tiny{$\pm 1.61$} \\  
       &w/ Fixed $\theta_{rps}$+CA  & IN1K-Self & 81.83\tiny{$\pm 0.12$} & 55.59\tiny{$\pm 0.21$} & 70.67\tiny{$\pm 0.02$} & 57.01\tiny{$\pm 0.07$} \\ 
       &w/ Fixed $\theta_{rps}$+CA+LN & IN1K-Self & 81.95\tiny{$\pm 0.17$} & 56.47\tiny{$\pm 0.23$} & 72.97\tiny{$\pm 0.17$} & 63.00\tiny{$\pm 0.21$} \\       
       &w/ SL+CA &IN1K-Self & 84.64\tiny{$\pm 0.21$} & 67.54\tiny{$\pm 0.29$} & 72.52\tiny{$\pm 0.06$} & 64.80\tiny{$\pm 0.20$} \\    
       &w/ SL+CA+LN &IN1K-Self &\textbf{85.27}\tiny{$\pm 0.08$}  &\textbf{68.07}\tiny{$\pm 0.21$}  & \textbf{73.01}\tiny{$\pm 0.16$}  & \textbf{66.04}\tiny{$\pm 0.08$} \\
       \hline
	\end{tabular}
	}
}
	\label{table:ablation}
    \vspace{+0.1cm}
    \caption{Ablation study. Here we present the Last-Acc (\%) after continual learning of all classes.
    $^{\dag}$The reproduced performance of Seq FT is slightly different from the reported one in~\cite{wang2022l2p,wang2022dualprompt} due to the use of different random seeds. SL: Slow Learner; LN: Logit Normalization; CA: a naive implementation of Classifier Alignment without LN.} 
\end{table*}

\textbf{Ablation Study:} We present an extensive ablation study of our approach in Table.~\ref{table:ablation}. 
To demonstrate the necessity of the proposed Slow Learner (SL), we consider two baselines: (1) sequential fine-tuning (Seq FT), which adopts a uniform learning rate of 0.005; and (2) Seq FT with fixed $\theta_{rps}$, which keeps the representation layer fixed and continually adjusts the classification layer.
Seq FT with fixed $\theta_{rps}$ is generally superior to Seq FT while significantly inferior to Seq FT w/ SL, indicating the \emph{necessity} of updating the representation layer (but with a properly reduced learning rate to mitigate the progressive overfitting problem). 

We further validate the effectiveness of the proposed Classifier Alignment (CA) and its Logit Normalization (LN). 
In particular, as the fine-grained continual learning benchmarks severely exacerbate the problem of classification layer (Fig.~\ref{img21k-cifar100-probe}), the benefits of SL are still significant but slightly reduced (\textit{e.g.}, for ImageNet-21K supervised pre-training, the improvements of SL are \textbf{47.09\%}, \textbf{44.85\%}, \textbf{28.05\%} and \textbf{22.17\%} on Split~CIFAR-100, Split~ImageNet-R, Split~CUB-200 and Split~Cars-196, respectively), while the benefits of CA+LN are greatly enhanced (\textit{e.g.}, the improvements of CA+LN are \textbf{2.67\%}, \textbf{5.20\%}, \textbf{19.64\%} and \textbf{17.99\%} on Split~CIFAR-100, Split~ImageNet-R, Split~CUB-200 and Split~Cars-196, respectively for IN21K-Sup pre-train). We also evaluate CA with fixed representations $\theta_{rps}$, which gain consistent improvements as well. Please note that our CA or CA+LN is operated in a \emph{post-hoc} fashion rather than aligning the classifier during representation learning~\cite{zhu2021pass,hayes2020remind,hayes2020lifelong}, as the latter would even \emph{worsen} the performance (\eg, by \textbf{27.89\%} on Split~CIFAR-100).

\textbf{Pre-training Paradigm and Downstream Granularity:} Here we analyze two critical factors for CLPM identified in our results.
Compared to supervised pre-training, self-supervised pre-training usually results in larger performance gaps between continual learning baselines and joint training (Fig.~\ref{img1k-cifar100-pt}, Table~\ref{table:cifar+inr}, \ref{table:cub+cars}). Although our proposal can greatly improve the performance of continual learning, the effectiveness varies with the choice of self-supervised methods (Fig.~\ref{img1k-cifar100-pt}). This is because they differ in the magnitude of updates required to learn all tasks well (Fig.~\ref{ssl_cka_jt_seq}), resulting in different degrees of difficulty in refining the representation layer.
Given that the large amount of data required for pre-training is typically unlabeled and incrementally collected, we suggest subsequent works to develop self-supervised pre-training paradigms that are more suitable for downstream continual learning, and potentially use this as a criterion to evaluate the progress of self-supervised learning.
On the other hand, the performance gap is increased as downstream continual learning becomes more fine-grained (Table~\ref{table:cifar+inr}, \ref{table:cub+cars}, \ref{table:ablation}). This is primarily due to the sub-optimal classification layer (Fig.~\ref{img21k-cifar100-probe}), which can be greatly improved by our classifier alignment strategy (Table~\ref{table:ablation}).


\textbf{Scalability:} We further discuss the scalability of our method.
First, the generated features are only used to align the output layer at test time rather than training the entire backbone, thus the \emph{computation} is efficient and not accumulated in continual learning (e.g., ranging from only 0.67\% to 5\% of the total running time for various benchmarks).
Second, the covariance $\Sigma_c$ can be further simplified as the variance $\sigma_c^2 \in \mathbb{R}^{d}$ with tolerable performance degradation (e.g., up to only 0.61\% on Split CIFAR-100 and 0.83\% on Split ImageNet-R). In this way, the \emph{storage} of $\mu_c$ and $\sigma_c^2$ for 100 classes corresponds to only 0.18\% parameters of the ViT-B/16 backbone, which is clearly lightweight. 

\section{Conclusion}
Continual learning on a pre-trained model (CLPM) requires effective transfer of pre-trained knowledge to each incremental task while maintaining its generalizability for future tasks. However, using a uniform learning rate to update the entire model makes traditional continual learning baselines fail to accommodate both objectives. 
Observing that using a slow learner can almost address this challenging issue in the representation layer, we further improve the classification layer through a classifier alignment strategy. Across a variety of upstream and downstream scenarios, our proposal enables the simplest sequential fine-tuning baseline to almost reach the joint training upper bound, far beyond the current state-of-the-art performance. 
Such a simple but extremely effective approach provides a strong criterion to \emph{re-evaluate the current progress and technical route} for CLPM in CV.
In addition, our empirical study demonstrates two critical factors for CLPM, such as pre-training paradigm and downstream granularity. Subsequent works could further explore these directions and develop more powerful strategies based on our proposal, so as to better leverage pre-trained knowledge for continual learning.

\textbf{Discussion of Limitations.}
Although we argue that self-supervised pre-training enjoys an advantage of less forgetting in upstream continual learning and propose an effective approach to overcome its shortcomings in downstream continual learning, how to perform upstream and downstream continual learning together remains to be explored. 
Besides, our results are based on a pre-trained ViT backbone for a fair comparison with previous works \cite{wang2022l2p,wang2022dualprompt}, without considering other representative architectures such as ResNet and other CL applications on downstream computer vision tasks such as object detection~\cite{shmelkov2017incremental}, image segmentation~\cite{MicroSeg,zhang2023coinseg,zhang2021few}. 
We leave the exploration of our work in the mentioned aspects to future works.




{\small
\bibliographystyle{ieee_fullname}
\bibliography{egpaper_final.bib}
}


\appendix

\section{More Details and Results.}
\textbf{Implementation Details.} All baselines follow an implementation similar to the one described in \cite{wang2022l2p,wang2022dualprompt}. Specifically, we adopt a pre-trained ViT-B/16 backbone. We use an Adam optimizer for prompting-based approaches that keep the representation layer fixed, while a SGD optimizer for other baselines that update the entire model, with the same batch size of 128. The original implementation of \cite{wang2022l2p,wang2022dualprompt} adopts a constant learning rate of 0.005 for all baselines, while our slow learner using 0.0001 for the representation layer and 0.01 for the classification layer. In practice, we observe that supervised pre-training usually converges faster than self-supervised pre-training in downstream continual learning. Therefore, for supervised pre-training, we train all baselines for 20 epochs on Split CIFAR-100 and 50 epochs on other benchmarks. For self-supervised pre-training, we train all baselines for 90 epochs on all benchmarks.

\begin{figure*}[h]
\hsize=\textwidth
    \centering
    \begin{minipage}[t!]{0.45\textwidth}
    \centering
    \includegraphics[width=7cm]{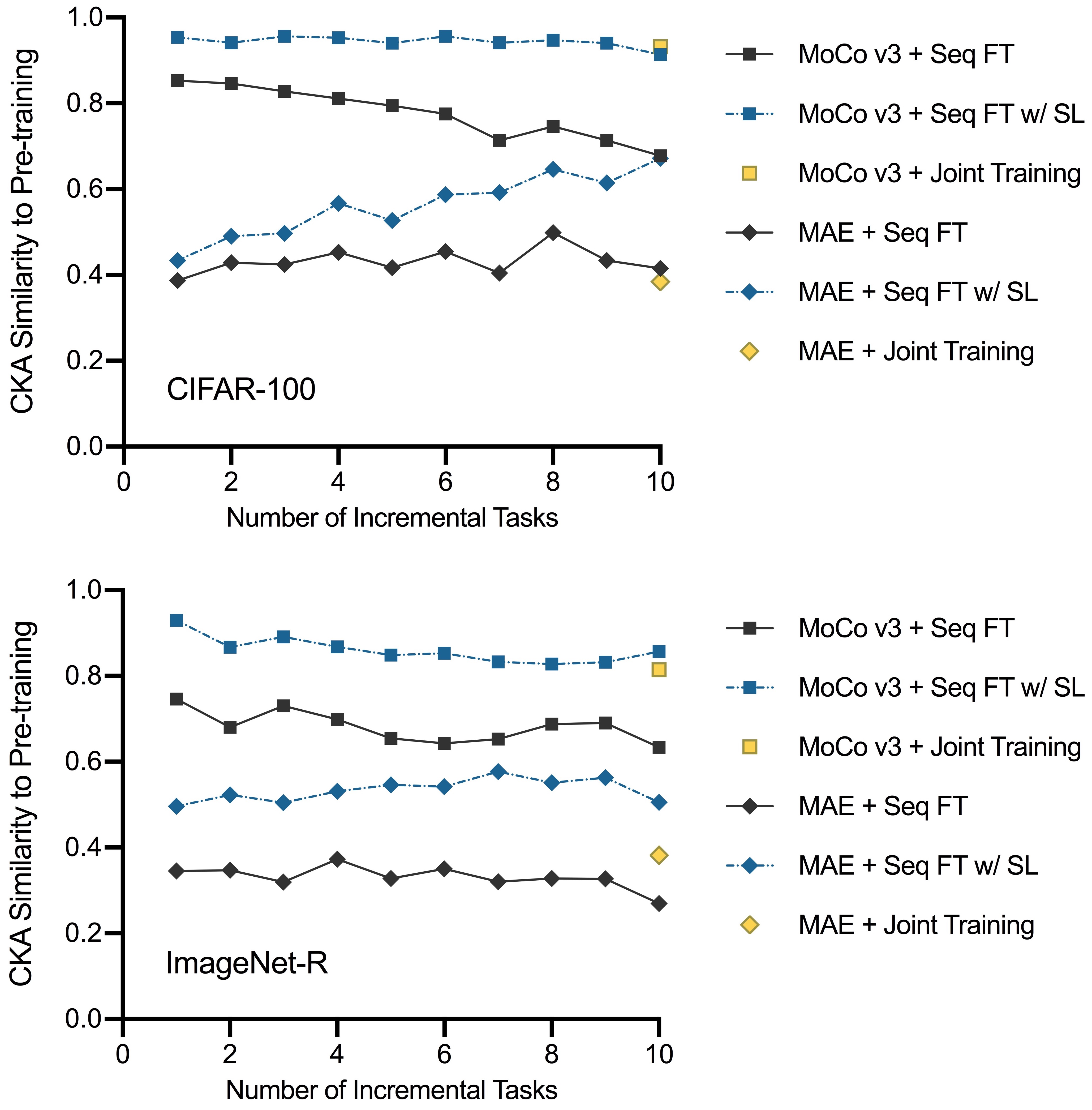}
    \caption{CKA similarity of pre-trained representations before and after learning downstream tasks.
    }
    \label{ssl_cka_cl_pt}
    \end{minipage}
    \hskip1em
    \centering
    \begin{minipage}[t!]{0.45\textwidth}
    \centering
    \includegraphics[width=7cm]{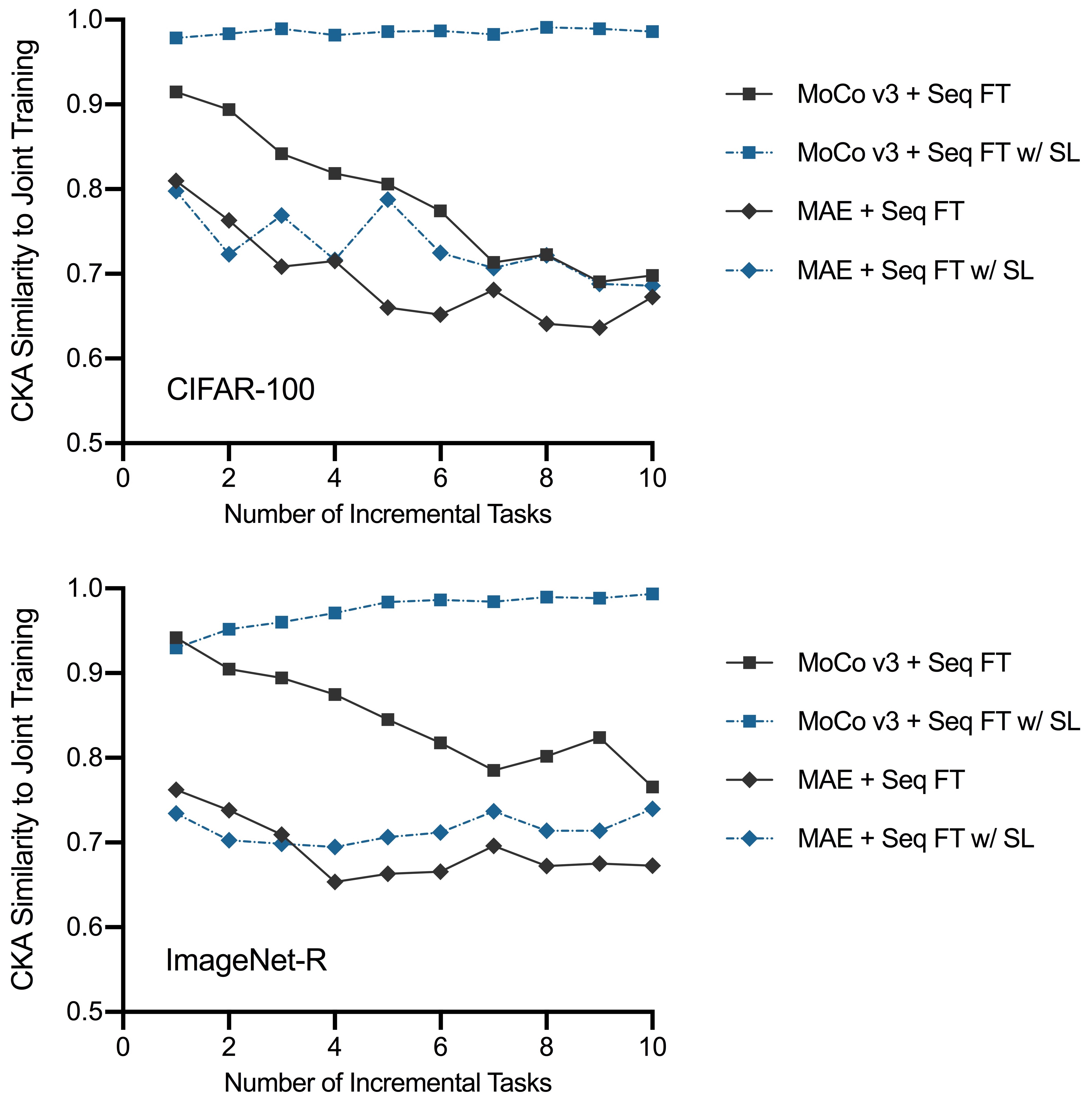}
    \caption{CKA similarity of pre-trained representations after joint training and after continual learning.
    }
    \label{ssl_cka_cl_jt}
    \end{minipage}
\end{figure*}

\textbf{Extended Analysis.} 
In this section, we provide extended results to support the main claims in our paper. First, we present the CKA similarity of pre-trained representation (1) before and after learning downstream tasks in Fig.~\ref{ssl_cka_cl_pt}, and (2) after joint training and after continual learning in Fig.~\ref{ssl_cka_cl_jt}.

\textbf{Results on Additional Dataset.} Except CIFAR-100, CUB-200-2011, ImageNet-R and Cars-196, we further consider a subset of DomainNet with 345-class sketch images (for short, Sketch-345). 
Our SLCA delivers consistently strong performance as shown in Table~\ref{sketch345}.

\textbf{Combine with other methods.} 
In the main text, the efficacy of SL has been widely validated by combining it with all baseline methods. We have further validated the efficacy of CA, presenting representative non-replay and replay methods on IN21K-Sup as shown in Table~\ref{table:combine}.

\begin{table}[h]
\smallskip
\centering
\setlength{\tabcolsep}{0.9mm}{
\resizebox{0.47\textwidth}{!}{
\begin{tabular}{l|cc|cc}
\hline
 & \multicolumn{2}{c|}{Sketch-345, IN21K-Sup} & \multicolumn{2}{c}{Sketch-345, IN1K-Self} \\ 
Method & Last-Acc (\%) & Inc-Acc (\%) & Last-Acc (\%) & Inc-Acc (\%) \\ \hline
Joint-Training & 72.18\tiny{$\pm 0.03$}  & - & 66.04\tiny{$\pm 0.07$}  & - \\ 
\hline
Seq~FT & 40.40\tiny{$\pm 14.87$} & 46.91\tiny{$\pm 24.25$} & 12.98\tiny{$\pm 4.09$} & 38.80\tiny{$\pm 5.49$}  \\
w/ SL & 63.41\tiny{$\pm 0.53$} & 71.24\tiny{$\pm 0.67$} & 56.94\tiny{$\pm 0.05$} & 66.07\tiny{$\pm 0.38$} \\
w/ SL+CA & \textbf{64.92}\tiny{$\pm 0.81$}  &\textbf{72.69}\tiny{$\pm 0.57$} & \textbf{59.88}\tiny{$\pm 0.06$}  &\textbf{67.99}\tiny{$\pm 0.54$} \\ 
 \hline
\end{tabular}\label{table:sketch}}}
\caption{Results on Sketch-345, a subset of DomainNet dataset~\cite{peng2019moment}. \label{sketch345} }
\end{table}

\begin{table}[h]
\smallskip
\renewcommand\arraystretch{0.95}
\resizebox{0.48\textwidth}{!}{
\begin{tabular}{lcccc}
\hline
Method & CIFAR-100 & ImageNet-R & CUB-200 & Cars-196 \\ \hline
EWC & 47.01\tiny{$\pm 0.29$} & 35.00\tiny{$\pm 0.43$} & 51.28\tiny{$\pm 2.37$} & 47.02\tiny{$\pm 3.90$} \\
EWC w/ SL &  89.30\tiny{$\pm 0.23$} & 70.27\tiny{$\pm 1.99$} & 81.62\tiny{$\pm 0.34$} & 64.50\tiny{$\pm 0.36$} \\ 
EWC w/ SL+CA &  \textbf{90.61}\tiny{$\pm 0.17$} & \textbf{71.48}\tiny{$\pm 0.31$} & \textbf{84.29}\tiny{$\pm 0.37$} & \textbf{69.61}\tiny{$\pm 0.29$} \\ 
\hline
BiC & 66.11\tiny{$\pm 1.76$} & 52.14\tiny{$\pm 1.08$} & 78.69\tiny{$\pm 1.97$} & 55.03\tiny{$\pm 3.27$} \\
BiC w/ SL & 88.45\tiny{$\pm 0.57$} & 64.89\tiny{$\pm 0.80$} & 81.91\tiny{$\pm 2.59$} &  63.10\tiny{$\pm 5.71$} \\
BiC w/ SL+CA & \textbf{91.57}\tiny{$\pm 0.13$} & \textbf{74.49}\tiny{$\pm 0.08$} & \textbf{86.82}\tiny{$\pm 0.69$} &  \textbf{73.90}\tiny{$\pm 0.38$} \\
\hline
\end{tabular}}
\caption{Ablations for CA combining with EWC and BiC. \label{table:combine}}
\end{table}

\begin{table*}[ht]
\hsize=\textwidth
   \centering
	\smallskip
 { 
	\begin{tabular}{c|c|c|c|c|c|c|c}
	 \hline
       Benchmark & Pre-trained & 0.005$^{\dag}$ & 0.001 & 0.0001 & 0.00001 & 0.000001 & Fixed $\theta_{rps}$\\
        \hline
        Split CIFAR-100 &IN21K-Sup &44.77 \small{$\pm 13.8$} &83.04\small{$\pm 1.46$} &88.86\small{$\pm 0.83$} &88.81\small{$\pm 0.46$} &85.11\small{$\pm 0.42$} &63.75\small{$\pm 0.67$}\\ 
        Split ImageNet-R &IN21K-Sup &26.95 \small{$\pm 11.8$} &70.38\small{$\pm 0.80$} &71.80\small{$\pm 1.45$} &62.64\small{$\pm 2.35$} &53.57\small{$\pm 4.33$} &34.64\small{$\pm 14.3$}\\  
        Split CUB-200 &IN21K-Sup &40.02 \small{$\pm 1.08$} &60.02\small{$\pm 1.24$} &68.07\small{$\pm 1.09$} &66.58\small{$\pm 3.93$} &64.38\small{$\pm 3.36$} &60.44\small{$\pm 1.80$}\\  
        Split Cars-196 &IN21K-Sup &27.57 \small{$\pm 1.79$} &15.74\small{$\pm 26.3$} &49.74\small{$\pm 1.25$} &30.66\small{$\pm 9.01$} &24.85\small{$\pm 7.90$} &24.51\small{$\pm 6.90$}\\  
        \hline
        \hline
        Split CIFAR-100 &IN1K-Self &27.99 \small{$\pm 5.16$} &81.49\small{$\pm 0.75$} &81.47\small{$\pm 0.55$} &81.57\small{$\pm 0.14$} &78.61\small{$\pm 0.29$} &77.30\small{$\pm 0.56$}\\ 
        Split ImageNet-R &IN1K-Self &45.84 \small{$\pm 4.19$} &68.72\small{$\pm 0.48$} &64.43\small{$\pm 0.44$} &59.19\small{$\pm 0.33$} &54.54\small{$\pm 0.32$} &51.97\small{$\pm 0.17$}\\  
        Split CUB-200 &IN1K-Self &45.35 \small{$\pm 1.38$} &68.58\small{$\pm 1.16$} &61.67\small{$\pm 1.37$} &56.46\small{$\pm 1.86$} &55.10\small{$\pm 2.13$} &55.54\small{$\pm 1.55$}\\  
        Split Cars-196 &IN1K-Self &35.96 \small{$\pm 2.04$} &58.39\small{$\pm 2.31$} &52.91\small{$\pm 1.61$} &43.64\small{$\pm 0.73$} &41.74\small{$\pm 0.23$} &43.16\small{$\pm 0.12$}\\  
        \hline
	\end{tabular}
	}
	\label{table:learning_rate}
      \vspace{+0.1cm}
	\caption{Continual learning performance with different learning rates of the representation layer. Here we present the Last-Acc (\%) after continual learning of all classes. IN21K-Sup: supervised pre-training on ImageNet-21K. IN1K-Self: self-supervised pre-training on ImageNet-1K with MoCo v3 \cite{chen2021empirical}. The column labeled by $^{\dag}$ uses the same learning rate of 0.005 for the entire model, while the others use a learning rate of 0.01 for the classification layer.} 
\end{table*}

\end{document}